\documentclass{article}

\usepackage[final]{nips_2016}
\usepackage[utf8]{inputenc} 
\usepackage[T1]{fontenc}    
\usepackage{hyperref}       
\usepackage{url}            
\usepackage{booktabs}       
\usepackage{amsfonts}       
\usepackage{nicefrac}       
\usepackage{microtype}      
\usepackage{graphicx}       

\title{Computing threshold functions using dendrites}

\author{
  Romain D. Caz\'{e} \\
  European Institute for Theoretical Neuroscience \\
  Centre Nationale de la Recherche Scientifique \\
  Paris, France \\
  \texttt{romain.caze@gmail.com} \\
  \And
  Bartosz Tele\'{n}czuk \\
  European Institute for Theoretical Neuroscience \\
  Centre Nationale de la Recherche Scientifique \\
  Paris, France \\
  \And
  Alain Destexhe \\
  European Institute for Theoretical Neuroscience \\
  Centre Nationale de la Recherche Scientifique \\
  Paris, France \\
}

\begin{document}

\maketitle

\begin{abstract}
Neurons, modeled as linear threshold unit (LTU), can in theory compute all threshold functions. In practice, however, some of these functions require synaptic weights of arbitrary large precision. We show here that dendrites can alleviate this requirement. We introduce here the non-Linear Threshold Unit (nLTU) that integrates synaptic input sub-linearly within distinct subunits to take into account local saturation in dendrites. We systematically search parameter space of the nTLU and TLU to compare them. Firstly, this shows that the nLTU can compute all threshold functions with smaller precision weights than the LTU. Secondly, we show that a nLTU can compute significantly more functions than a LTU when an input can only make a single synapse. This work paves the way for a new generation of network made of nLTU with binary synapses. 
  
\end{abstract}
\section{Introduction}
Neurons constitute the basic computational unit in the brain but they are also complex structures that can implement various computations \citep{Koch2000}. Neurons receive their inputs on dendrites that form elegant tree-like structures. These inputs can non-linearly interact: the response to two inputs can either be lower or higher than their arithmetic sum \citep{Abrahamsson2012, Polsky2004, Koch1983}. These non-linear interactions enhance the computing capacity of a single neuron and could turn it into a two layers neural network \citep{Poirazi2003a}. Moreover, even sublinear interactions can enhance the computational capacity of neurons \citep{Caze2013}.   

Linear Threshold Unit (LTU) has been the standard model to describe how a single neuron might compute \citep{Rosenblatt1958}. This model sums linearly its input using weights and the resulting sum determines its activity, i.e. if this sum crosses a threshold the unit is active and otherwise inactive. The LTU can potentially compute all threshold functions that are linearly separable \citep{Minsky1957}. To reach its maximum computational capacity, however, the model might need weights to use finely tuned synaptic weights. Synaptic weights have a limited precision in practice and some even considered that brain works with  binary-valued synapses. In this case the LTU cannot compute all threshold functions \citep{Draghici2002}. We can then wonder what are their real computing capacity?

We introduce here the non-Linear Threshold Unit (nLTU). We are going to compare this model with the LTU using limited precision weights. The nLTU features multiple units that can saturate at a given threshold; the outputs of these units are summed and passed though a Heaviside step function to obtain the model output (see Fig.~\ref{fig1} {\it Right}). These multiple units account for the local and non-linear interactions in dendrites. Contrary to the LTU, it is possible to use the position of synapses -- on which subunit they connect -- to implement function. We suspect that this capacity will extend the computing capacity for the same range of synaptic weight.

\section{Using systematic parameter search to measure the computing capacity}
We wanted to determine how many functions a model could compute within a given parameter range. Therefore, we determined this computing capacity using an exhaustive parameter search. For all the parameters we used an integer valued range. And we stopped the search because of two distinct conditions:
\begin{enumerate}
\item The maximal computing capacity for the LTU is reached, i.e we reached the maximal number of threshold function.
\item We limited the search to a given number of synapses per input
\end{enumerate}
For each parameter set corresponds a unique function. Therefore, by exhaustively searching the integer parameter range, we are sure to find all functions computed using this parameter range. For instance, we demonstrate in Fig.~\ref{fig1} how a LTU or a nLTU can compute the same function of three inputs.

\section{Computing the same threshold function with a LTU or a nLTU}

\begin{figure}[h]
\begin{center}
  \includegraphics[scale=0.5]{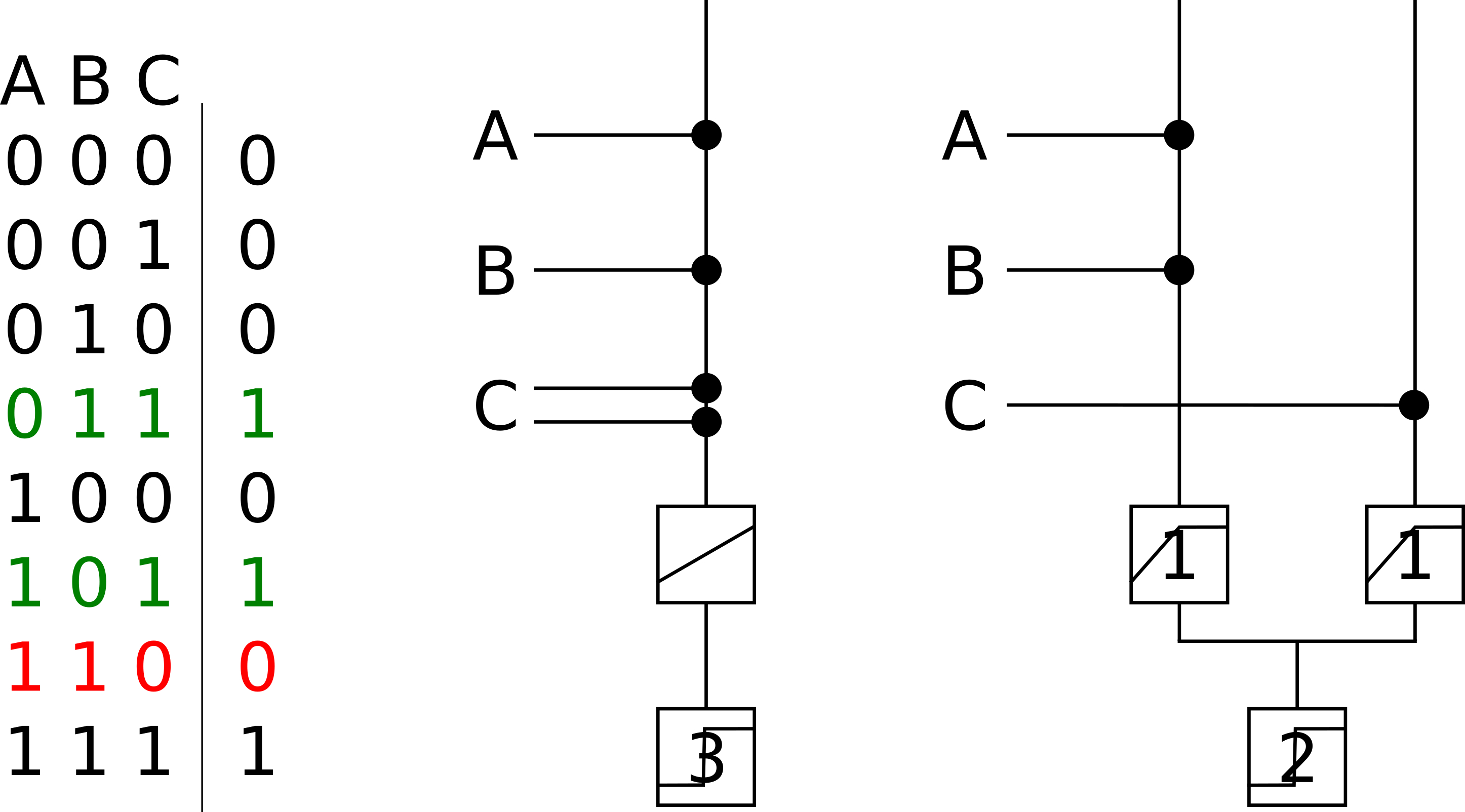}
\end{center}
  \caption{Schematic representation of a threshold Linear Unit (LTU, middle), a threshold non-Linear threshold subunits (TLUnm, middle) and the function they compute. Circles stand for synapses and squares depict linear or non-linear filters (saturating piece-wise linear or Heaviside function). Numbers in squares are either the saturation or the threshold of Heaviside function. The truth table on the right describes the input output function of these artificial neurons}
  \label{fig1}
\end{figure}

Our parameter search enabled us to find that the same function can be implemented differently by a LTU or nLTU Fig.~\ref{fig1}. This function is the simplest that could distinguish LTU from a nLTU. A neuron implementing this function should respond when the population AC or BC are co-active but not when AB is active Fig.~\ref{fig1} {\it Left}. Implementing this function requires to give an edge to the ensemble C. In the case of the LTU, it is achieved using strength. The ensemble C is making more synaptic contact than A or B, Fig.~\ref{fig1} {\it middle}. In the case of the nLTU, one can implement the function using a different method. The neuron will not respond to AB because A and B both make synapse on the same dendritic subunit, Fig.~\ref{fig1} {\it right}. For this reason they will interact sub-linearly. But the neuron will fire when AC or BC activate because the two couples of inputs interact linearly. In summary, the LTU will need at least two synapses coming from the ensemble C whereas for a nLTU a single synapse suffices. 

\section{Reaching the capacity of the LTU with small precision weights}

\begin{figure}[h]
\begin{center}
  \includegraphics[scale=0.8]{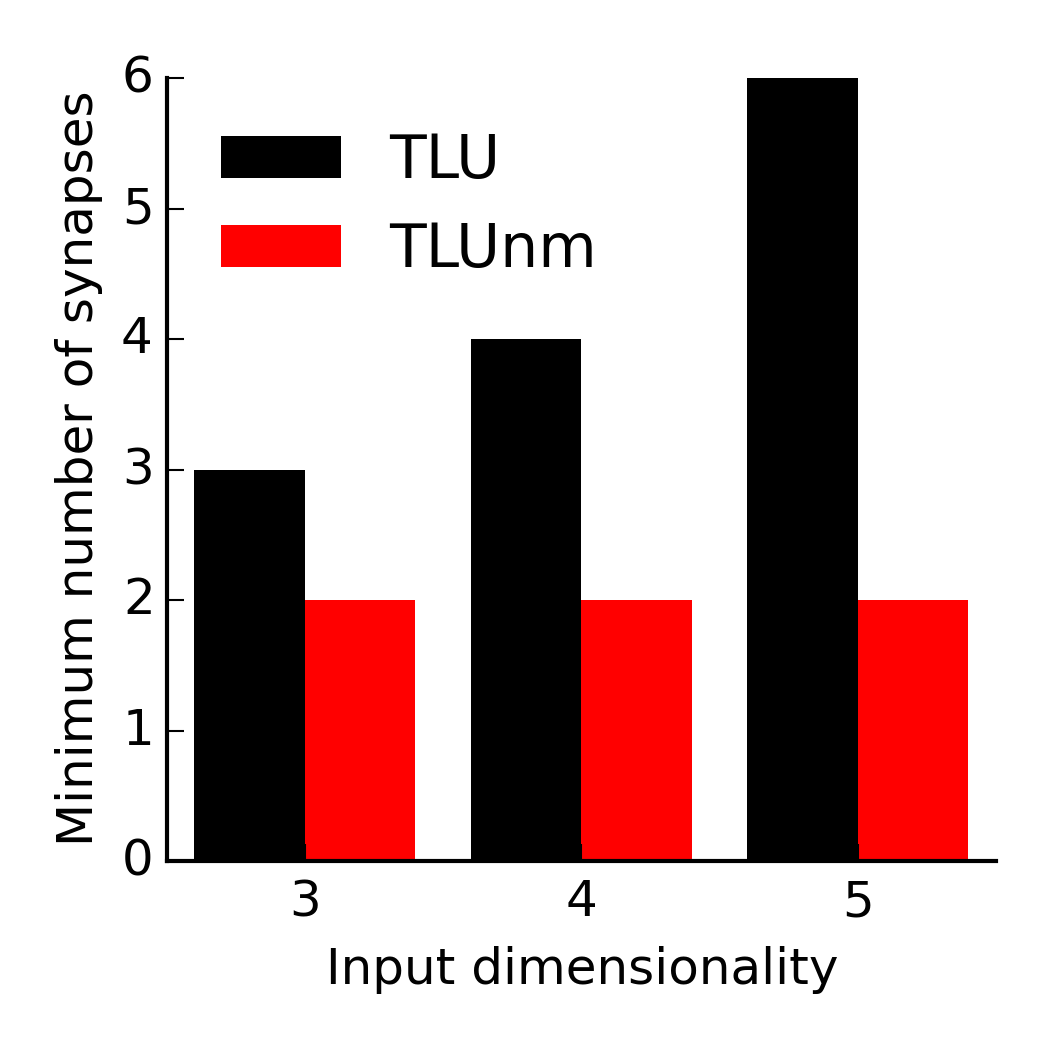}
\end{center}
\caption{Number of synapses necessary to reach the maximal capacity of the LTU (black) for given number of inputs (input dimensionality abscissa). In red the number of synapses required to reach the same capacity with a non-linear threshold unit (nLTU). }
  \label{fig2}
\end{figure}

The difference between LTU and nLTU described in the previous section is small for a low number of inputs. How does this difference evolve with the number of inputs? We used a systematic parameter search to determine what is happening in the case of higher number of inputs. The results for three, four or five inputs are presented on Fig.~\ref{fig2}. Surprisingly, two synapses per input line are sufficient for the nLTU to compute all the linearly separable functions, which can be potentially implemented by the LTU, whereas a LTU requires respectively three, four or six synapses per input to reach the same capacity. Studying the case of six inputs becomes very intensive but this might be possible to calculate it with existing computers. The case of five inputs, however, displays already a 3-fold difference between the LTU and nLTU. We expect this difference to grow exponentially. One could explain this difference by looking at how a function can be implemented by the nLTU (see Fig.\ref{fig1}). While for a LTU it is only possible to use strength, to play on the number of synapses per input, for a nLTU it is also possible to use space, i.e. placing the synapses on distinct subunits. Using this former method enables to make tremendous economy on the number of synapses required to implement a function. 

\section{A higher capacity for nLTU than LTU with a unique synapse per input line}
\begin{figure}[h]
\begin{center}
  \includegraphics[scale=0.8]{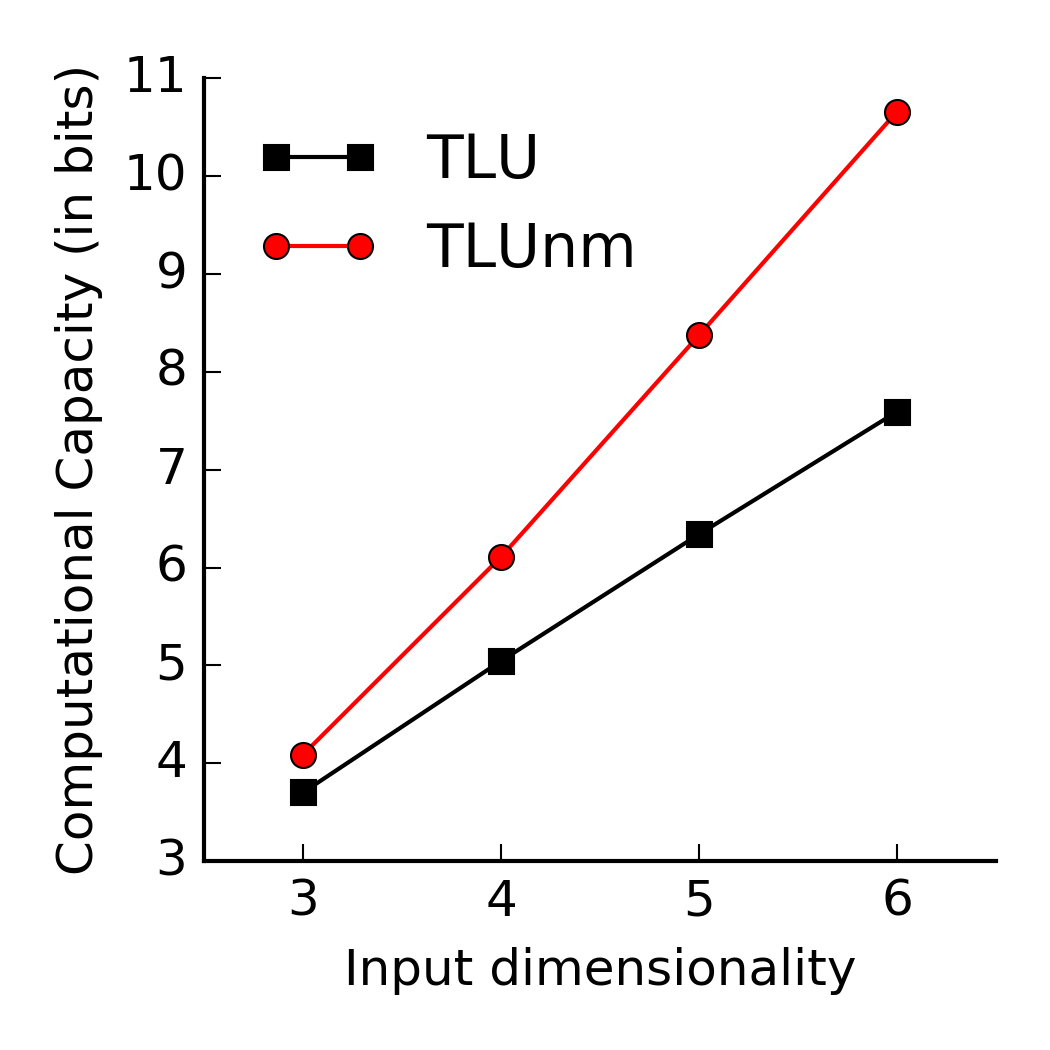}
\end{center}  
\caption{The capacity (base-2 logarithm of the number of computable functions) by the LTU (black squares) or by the non-linear unit (red points, nLTU). The capacity depends on the number of inputs, i.e. how many binary digits are required to represent all inputs.}
  \label{fig3}
\end{figure}

What happens when each input line has only one synapse? We studied this situation both for the LTU and for the nLTU. We manage to go up to six inputs where the gap between the LTU and nLTU is large (see Fig.~\ref{fig3}). The gap between the LTU and nLTU grows linearly with the number of inputs. For instance, for five inputs the number of functions that can be computed by a nLTU is 332 while it is only 81 for a LTU. This results in a capacity of approximatively 6 bits for nLTU and 8 bits for the LTU. For six inputs the difference increase to 1000 for the nTLU and only 128 for the TLU meaning an order of magnitude higher for the nTLU.

\section{Conclusion/Discussion}
We have demonstrated here that the threshold non-linear muti-units (TLUnm) can compute linearly separable functions with small precision weights. While a threshold function requires up to two synapse per input line, the TLUnm only requires at most one synapse per input line (see Fig.~\ref{fig1}). This difference could seem negligible but it grows exponentially with the number of input lines --the input dimension --  (see Fig.~\ref{fig2}). Even for a single synapse per input line and a low input dimension the computational capacity of the TLUnm outperform the one of a TLU (see Fig.~\ref{fig3}).

The TLUnm might look like a two layer network of threshold linear unit but is not. They are different on three different levels: mathematical, because the unit standing for dendrites are purely sublinear while a Heaviside function is both sub and supralinear; computational, because these two objects do not have the same computational capacity \citep{Caze2012}; and biological, because dendritic spikes require active mechanism for spike regeneration (voltage-gated channels). In summary, a TLUnm differ substantially for a two layer network of TLUs and may be more energy efficient for the biological implementation. 

Our model might have a major impact on the field of neural networks which could be implemented in neuro-inspired analog chips \citep{Schemmel2010}. The known limitations of such chips is the large variability of circuit parameters defining the neuron, therefore the use of TLUnm with limited precision weights seem adequate for implementation on this hardware.

\section*{Supplementary Material}
A zip file called ``Scripts.zip'' enables to reproduce the second and third figure. Upon publication of the manuscript this code will be available on Github.

\bibliography{MyCollection}
\bibliographystyle{plainnat}

\end{document}